\pdfoutput=1

\documentclass[11pt]{article}

\usepackage[preprint]{acl}
\usepackage{amsmath}
\usepackage{amsfonts}
\usepackage{times}
\usepackage{tabularx}
\usepackage{booktabs}
\usepackage{multirow}
\usepackage{siunitx}
\usepackage{latexsym}
\usepackage{url}
\usepackage[mathscr]{eucal}
\usepackage{pifont} 
\usepackage{multirow}
\usepackage{adjustbox}
\usepackage{algorithmic}
\usepackage{algorithm}
\usepackage{hyperref}
\usepackage[normalem]{ulem}

\usepackage{todonotes}

\usepackage[T1]{fontenc}

\usepackage[utf8]{inputenc}

\usepackage{microtype}
\usepackage{booktabs}
\usepackage{inconsolata}

\usepackage{graphicx}

\title{Tree of Agents: Improving Long-Context Capabilities of Large Language Models through Multi-Perspective Reasoning}

\author{
\textbf{Song Yu\textsuperscript{1}}, 
\textbf{Xiaofei Xu\textsuperscript{2}}, 
\textbf{Ke Deng\textsuperscript{3}\textsuperscript{\dag}}, 
\textbf{Li Li\textsuperscript{1}\textsuperscript{\dag}}, 
\textbf{Lin Tian\textsuperscript{4}} 
\\
\textsuperscript{1}School of Computer and Information Science, Southwest University, Chongqing, China \\
\textsuperscript{2}School of Information Technology, Murdoch University, Perth, Australia \\
\textsuperscript{3}School of Computing Technologies, RMIT University, Melbourne, Australia \\
\textsuperscript{4}University of Technology Sydney, Sydney, Australia
\\
\texttt{yusong0929@email.swu.edu.cn}\qquad
\texttt{xiaofei.xu@murdoch.edu.au} \\ \texttt{ke.deng@rmit.edu.au} \qquad \texttt{lily@swu.edu.cn}\qquad  \texttt{Lin.Tian-3@uts.edu.au}
}

\begin{document}
\maketitle

\footnotetext{\textsuperscript{\dag}Corresponding authors.}
\begin{abstract}
Large language models (LLMs) face persistent challenges when handling long-context tasks, most notably the \textit{“lost in the middle”} issue, where information located in the middle of a long input tends to be underutilized. Some existing methods that reduce input have the risk of discarding key information, while others that extend context windows often lead to attention dispersion. To address these limitations, we propose \textit{Tree of Agents (TOA)}, a multi-agent reasoning framework that segments the input into chunks processed by independent agents. Each agent generates its local cognition, then agents dynamically exchange information for collaborative reasoning along tree-structured paths. TOA enables agents to probe different reasoning orders for multi-perspective understanding, effectively mitigating position bias and reducing hallucinations. To improve processing efficiency, we incorporate prefix-hash caching and adaptive pruning strategies, achieving significant performance improvements with comparable API overhead. Experiments show that TOA, powered by compact LLaMA3.1-8B, significantly outperforms multiple baselines and demonstrates comparable performance to the latest and much larger commercial models, such as Gemini1.5-pro, on various long-context tasks. Code is available at \url{https://github.com/Aireduce952/Tree-of-Agents}.
\end{abstract}

\begin{figure}[t]
  \includegraphics[scale=0.45]{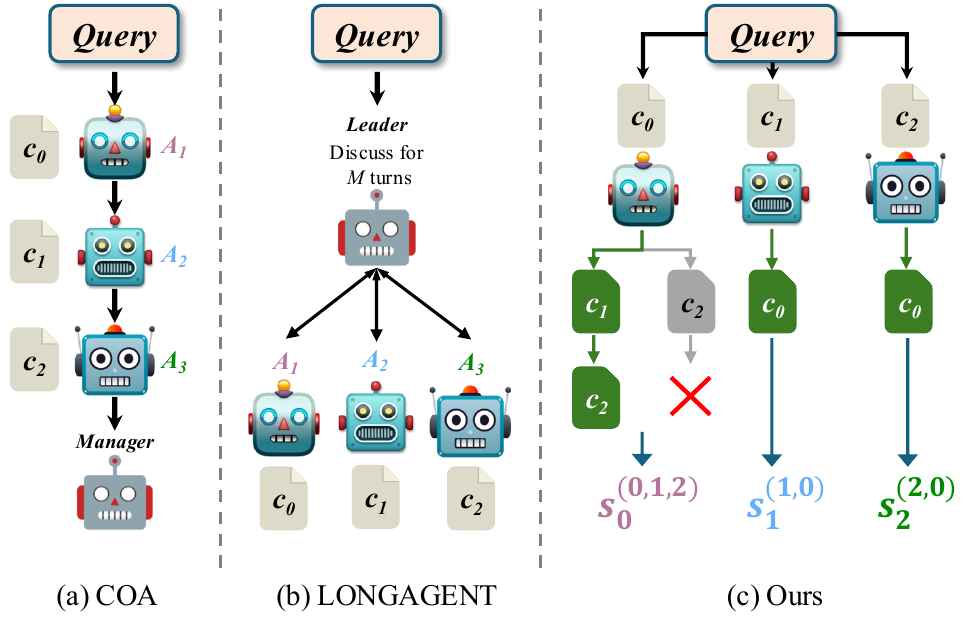} 
  \centering
  \caption {The core difference between TOA and other multi-agent reasoning methods.
(a) COA processes chunks sequentially, with a final manager agent making the decision.
(b) LONGAGENT uses a leader agent to coordinate multi-turn discussions with others.
(c) TOA probes multi-paths in a tree structure to prompt multi-perspective reasoning.}
  \label{fig:intro}
\end{figure}

\begin{table*}[h]
\small
\centering
\begin{tabular*}{\hsize}{@{}@{\extracolsep{\fill}}llccccc@{}}
\toprule
\textbf{Category}                        & \textbf{Example Work}           & \textbf{PnP} & \textbf{No Train}   & \textbf{Focus} & \textbf{Cost} & \textbf{Inter.}\\ \hline
\multirow{3}{*}{Model Modification} & Position Encoding \cite{zhu2024coca} & \ding{56}& \ding{56} & \ding{56}& High & Low    \\
                                    & Attention Mechanism \cite{chenlonglora}& \ding{56}& \ding{56} & \ding{56}& High&Low     \\
                                    & Training Strategies \cite{zhang2024longreward} & \ding{56} & \ding{56} & \ding{56}& High&Low     \\ \hline
\multirow{2}{*}{Input Reduction}    & RAG \cite{zhao2024longrag}   & \ding{56}     & \ding{56} & \ding{52}  & Medium&Medium     \\
                                    & Prompt Compression \cite{jiang2024longllmlingua} & \ding{56} & \ding{52} & \ding{52}  & Medium&Low   \\ \hline
\multirow{3}{*}{Multi-Agent Reasoning}    & LONGAGENT \cite{xiaoinfllm}   & \ding{56}     & \ding{56} & \ding{52}   & High&High     \\
& COA \cite{zhang2024chain} & \ding{52} & \ding{52} & \ding{52} &High&High \\

                                    & Tree of Agents (Ours)  & \ding{52} & \ding{52} & \ding{52}   & High&High   \\ \toprule
\end{tabular*}
\caption{Comparison of various existing methods in long context tasks. Plug-and-play(PnP): whether the method can be applied directly without additional adaptation. Focus: ability to identify and prioritize critical information in long contexts. Inter.: interpretability of each method.}
\label{tbl:existing_approach}
\end{table*}
\section{Introduction}
The capabilities of Large Language Models (LLMs) have seen rapid improvements in recent years. The advent of models such as OpenAI-o3~\cite{el2025competitive} and DeepSeek-R1~\cite{guo2025deepseek} marks further progression in model inference capacity. Although LLMs perform well in many scenarios, they still fall short in tasks involving long contexts, such as financial report comprehension~\cite{reddy2024docfinqa}, novel quizzes~\cite{wangnovelqa, xu2024detectiveqa} and legal contract analysis~\cite{shen2022multi, shu2024lawllm}. 
There are three primary challenges in long-context tasks: (1) Position biases processing further impair performance, often referred to as the "lost in the middle" issue~\cite{liu2024lost}. (2) As input length increases, the presence of extraneous or redundant information grows, potentially diluting the model’s focus and degrading output quality~\cite{shi2023large}. (3) The computational demands for both training and inference escalate dramatically with longer sequences~\cite{gao2024train}.

To address these issues, existing approaches can be categorized into three main directions, as shown in Table~\ref{tbl:existing_approach}. The first direction focuses on optimizing the model itself, either by enhancing long dependency capture through improvements in attention mechanisms~\cite{chenlonglora, yen2024long}, training strategies~\cite{zhang2024longreward, chen2024long}, or position encoding~\cite{zhu2024coca, jin2024llm}. The second direction aids the model in solving long contexts and reducing memory usage by enhancing key information density or reducing input length through retrieval augmented generation~\cite{zhao2024longrag,xu2023retrieval} or prompt compression~\cite{jiang2024longllmlingua}. 
The third direction addresses long text processing by employing multi-agent reasoning~\cite{zhang2024chain,zhao2024longagent} to break down the input into manageable chunks, which are subsequently integrated to arrive at the final result. 
Nonetheless, processing long contexts efficiently remains a challenging problem~\cite{hsiehruler}. 
Extending the context window for a model may hurt its performance on short texts~\cite{dinglongrope}. 
Retrieval-based method relies on the quality of external retrieval tools and may introduce noise~\cite{li2024retrieval}. 
Prompt compression may lose key information, especially when information is unevenly distributed in long texts. Such methods are more suitable for QA or information extraction tasks and have limited performance for tasks that require global understanding~\cite{jiang2023llmlingua}. 
Multi-agent reasoning requires a well-designed agent communication mechanism to get the right results. As illustrated in Figure \ref{fig:intro}, COA lets agents process text chunks sequentially and use a manager to integrate the results~\cite{zhang2024chain}, and LONGAGENT uses a leader agent to coordinate multi-turn discussions with others~\cite{xiaoinfllm}. However, these approaches generally do not support multi-perspective understanding of the document.
This study proposes a novel multi-perspective reasoning framework — \textbf{Tree of Agents} (TOA), inspired by cognitive science theories and empirical findings on problem-solving. As shown in Figure~\ref{fig:intro}, TOA splits a long text into chunks and employs multiple agents to explore different orders of chunks along multi-paths of a tree structure. This design draws from Kahneman’s dual-process theory~\cite{kahneman2011thinking}, which advocates for a multi-perspective approach, and Newell and Simon’s theory of problem-solving as an iterative exploration process~\cite{newell1959report}. Unlike sequential processing, which often struggles in tasks requiring strategic exploration or early decisions, our method disrupts the natural reading order to enable agents to re-assess and refine their understanding through cross-validation.

Agents in TOA collaborate by exchanging local cognition and applying the mechanism for consensus formation. The multi-perspective reasoning significantly improves performance on various tasks but increases computational cost. To mitigate this issue, we introduce two efficiency-enhancing strategies: 1) A prefix-hash-based cache mechanism that reduces redundant cognition generations; 2) An adaptive pruning strategy that terminates useless reasoning paths early.

We conduct extensive experiments with eight baselines on two long-context reasoning datasets and one benchmark, demonstrating the effectiveness of our approach. The main contributions are: 

1) We propose TOA, the first tree-structured multi-agent reasoning framework that addresses long-context modeling via multi-perspective reasoning. 

2) We introduce two optimization strategies that reduce the computational cost while maintaining the advantage of the proposed multi-perspective reasoning. 

3) This study empirically validates that TOA consistently improves long-context reasoning for various tasks. More interestingly, our experiments show that the different reading order of a long text can lead to different perceptions, and thus different answers to queries.  

\section{Related Work}

This section summarizes key directions in long-context modeling for LLMs. 

\subsection{Specialized Long-Context Models}

Training specialized models capable of handling long-contexts is a straightforward approach to improving capabilities. Notable examples include Claude3~\cite{TheC3} and Gemini-1.5pro~\cite{team2024gemini}, which support 200K and 2M context windows, respectively. While these models can process long texts more effectively, they require expensive computational resources and may reduce their ability to understand short texts~\cite{liu2024lost}. 

\subsection{Model Modification}

To reduce the computational cost of training specialized models, several methods based on positional encoding have been proposed to enhance long dependency capture. Such methods include constraining the covariance of Q/K vectors~\cite{zhu2024coca} or dynamically adjusting the RoPE angle~\cite{lin2024mixture}.
Additionally, SelfExtend expands the context window during the inference phase with only minor modifications to the model, without the need for fine-tuning~\cite{jin2024llm}. While effective in reducing training costs, these approaches often struggle with high complexity and computational limitations in practice. 

\subsection{RAG and Input Reduction}
LongRAG combines global and local retrieval to enhance the accuracy of long-context question answering tasks~\cite{zhao2024longrag}. The applicability of RAG with long-context models for downstream tasks has been analyzed by~\cite{xu2023retrieval}. These methods require additional information, which is not always available. Input reduction is another research direction that avoids modifying the model structure. In LongLLMLingua, the author compresses redundant information in the prompts to improve the density of key information~\cite{jiang2024longllmlingua}. However, input reduction has the risk of discarding key information.

\subsection{Hybrid Approaches}

Hybrid methods aim to dynamically select between retrieval-based and long-context models based on task complexity. Self-Route, for instance, adjusts the model's approach depending on the task at hand~\cite{li2024retrieval}. InfLLM stores long context externally and dynamically correlates related fragments during inference~\cite{xiaoinfllm}. These methods can be flexible and efficient, but the complexity of dynamically choosing between methods may introduce additional overhead. 

\subsection{Multi-agent Reasoning}

Recently, approaches based on multi-agent reasoning have gained attention as a promising solution to handle long contexts. HOMER reduces memory usage through chunking and token reduction~\cite{songhierarchical}, while COA allows agents to process text chunks sequentially and communicate to integrate results~\cite{zhang2024chain}. LONGAGENT coordinates agents through a leader agent to answer queries over 128K documents~\cite{zhao2024longagent}. COA and LONGAGENT avoid token reduction like HOMER. However, COA’s unidirectional messaging and lack of dynamic interactions can lead to broken inference chains. LONGAGENT’s performance depends on the leader's ability to decompose tasks, and fixed chunk sizes may disrupt paragraph integrity. In contrast, our TOA enables agents to probe different reasoning orders of chunks, effectively promoting multi-perspective understanding to mitigate position bias and reduce hallucinations. 

\begin{figure*}[t]
  \includegraphics[scale=0.4]{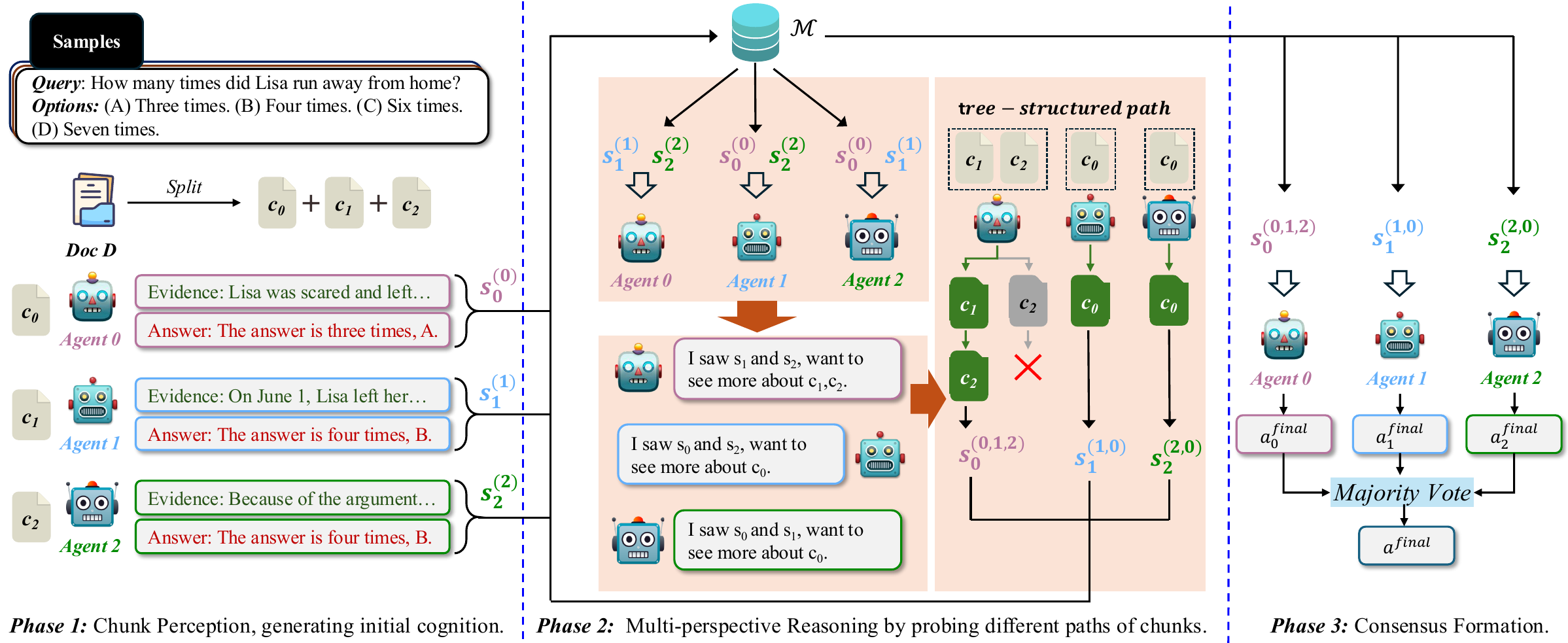} \hfil
  \centering
  \caption {An overview of TOA. In Phase 1, the document is split into chunks, with each agent processing a chunk and providing cognition, which are stored in $\mathcal{M}$. In Phase 2, agents exchange cognition, express interest in reading additional chunks, and probe additional chunks in different orders. 
  In Phase 3, each agent generates a local answer, and the final answer is determined by majority voting.}
  \label{fig:overview}
\end{figure*}

\section{Methodology}

We first formulate the problem that TOA aims to solve. Given a query $q$ and a document $\mathcal{D}=\{t_1, t_2,...,t_M\}$ with $M$ tokens, where $M$ is much longer than the context window length of the LLM. Our goal is to build a system that can respond to the query correctly according to the document. TOA follows a three-phase reasoning process, as shown in Figure \ref{fig:overview}. 

\subsection{Phase 1: Chunk Perception}
Due to the excessive context length, 
we first break down the document $\mathcal{D}$ into $N$ short text chunks $(c_0,\cdots,c_{N-1})$. For the $i$th chunk $c_i$, it consists of tokens from position $i \cdot\frac{M}{N}$ to $(i+1)\cdot\frac{M}{N}-1$, 
where $\frac{M}{N}$ is set to be below the agent's maximum context window. By adjusting the number of agents, TOA can handle varying input lengths, where one agent is for one chunk. For simplicity, the agent for the $i$th chunk is named agent $\mathcal{A}_i$.     

In phase 1, $\mathcal{A}_i$ generates the initial cognitive state based on the $i$th chunk. The input of $\mathcal{A}_i$ is $(q,c_i,p)$, where $q$ represents the user's query, $c_i$ represents the $i$th chunk, and $p$ is the prompt. $\mathcal{A}_i$ outputs $\langle e_i,a_i\rangle$, where $e_i$ is the evidence from chunk $c_i$ that support to answer query $q$, and $a_i$ is the answer returned by $\mathcal{A}_i$. We can express $\mathcal{A}_i$ as a function:
\begin{equation*}
    \langle e_i, a_i\rangle=\mathcal{A}_i(q,c_i,p), \forall i\in [0,N-1].
\end{equation*}
The initial cognitive state of agent $\mathcal{A}_i$ is denoted as $s_i^{(i)}=\langle e_i^{(i)}, a_i^{(i)}\rangle$ and stored in a buffer $\mathcal{M}_i$. In Figure \ref{fig:overview}, $\mathcal{M}$ denotes the buffers for all agents. 
To answer the question "How many times did Lisa run away from home?", each agent relies solely on its chunk. However, such multi-hop questions require a comprehensive understanding of the entire document. Lacking a global view, the answer of each agent is typically biased. 

\subsection{Phase 2: Multi-Perspective Understanding}
In phase 2, each agent reads the initial cognitive states of other agents stored in $\mathcal{M}$. As shown in Figure \ref{fig:overview} as an example, $\mathcal{A}_0$ reads $s_1^{(1)}$ and $s_2^{(2)}$, the initial cognitive states of $\mathcal{A}_1$ and $\mathcal{A}_2$, respectively. $\mathcal{A}_1$ and $\mathcal{A}_2$ do the same. The initial cognitive state of $\mathcal{A}_i$ can be viewed as a summary of the $i$th chunk by $\mathcal{A}_i$. By reading the initial cognitive states of other agents, an agent, already having an understanding of its chunk, can decide which other chunks can help it answer the query. For $\mathcal{A}_i$, the identities of these additional chunks are stored in $\mathcal{G}_i$. In Figure \ref{fig:overview}, $\mathcal{A}_0$ wants to read additional chunk $c_1$ and $c_2$, $\mathcal{A}_1$ and $\mathcal{A}_2$ wants to see $c_0$, respectively.



Then, TOA probes different paths following a tree structure for a comprehensive multi-perspective understanding of the document $\mathcal{D}$. For agent \(\mathcal{A}_i\), the additional chunks to read are indexed in $\mathcal{G}_i$. Suppose \(\mathcal{G}_i = \{ j_0, j_1, ..., j_{k-1} \}\). We define \( \text{Perm}_i \) as all different paths visiting the chunks in \( \mathcal{G}_i \). For example, if \( \mathcal{G}_i = \{0, 1, 2\} \), then \(\text{Perm}_i=\) \(\{(0,1,2),\) \((0,2,1),\) \((1,0,2)\), \((1,2,0),\) \((2,1,0),\) \((2,0,1)\)\(\}\). If connected by a root, these paths are the paths form the root to leaves in a tree structure. Each path represents a unique way to read these chunks in \(\mathcal{G}_i\). The purpose of probing all possible paths is to explore different cognitive orders, helping to avoid biases caused by the fixed reading order and ensuring a thorough understanding. If one path misses a key clue, other paths may fill in the gap. This redundancy mechanism greatly enhances robustness for complex tasks. \textit{We illustrate an example in Appendix \ref{appendix:order} to illustrate that the different reading orders lead to different answers by the same agent.}

Suppose the current cognitive state of \(\mathcal{A}_i\) is $s_i^{(i)}$. After reading a chunk $c_j$ indexed in \(\mathcal{G}_i\), the cognitive state of \(\mathcal{A}_i\) is updated as: 
\begin{equation*}
    s_i^{(i,j)}\gets \langle e_i^{(i,j)}, a_i^{(i,{j})}\rangle=\mathcal{A}_i(s_i^{(i)},c_j).
\end{equation*}
As shown in Figure \ref{fig:overview}, \(\mathcal{A}_0\) reads chunk $c_1$ and $c_2$ so that the cognitive state of \(\mathcal{A}_0\) is updated from $s_0^{(0)}$ to $s_0^{(0,1,2)}$. Similarly, the cognitive state of \(\mathcal{A}_1\) and \(\mathcal{A}_2\) are updated from $s_1^{(1)}$ to $s_1^{(1,0)}$ and from $s_2^{(2)}$ to $s_2^{(2,0)}$, respectively. 

While each agent \(\mathcal{A}_i\) to have a multi-perspective understanding of the document to answer queries correctly, the number of paths grows with the number of chunks in \(\mathcal{G}_i\). To mitigate such a combinatorial explosion, we propose two optimization strategies.




\subsubsection*{State Caching in Tree Structure} 

As discussed above, all paths related to an agent form a tree structure, i.e., a path of chunks corresponds to a path from the root to a leaf in a tree structure. Given two paths of chunks, for example (0,1,2,4) and (0,1,2,3), they share a chunk sequence (0,1,2) and then fork to chunk 4 and chunk 3. Once the cognitive state for the chunk sequence (0,1,2) has been generated for the path (0,1,2,4), there is no need to generate it again for the path (0,1,2,3). 

For this purpose, we store and reuse previously generated intermediate states by $\mathcal{A}_i$ in $\mathcal{M}_i$ with prefix hashing.
Recall that $c_i$ is the initial chunk of $\mathcal{A}_i$ in phase 1. The cache $\mathcal{M}_i$ is initialized as follows:
\begin{equation*}
    \mathcal{M}_i(\langle c_i \rangle)\leftarrow{s_i^{(i)}}.
\end{equation*}
For the $j$th path of $\mathcal{A}_i$ from $\text{Perm}_i$, $p_{ij}=\langle c_i, c_{j_0}, c_{j_1}, ..., c_{j_{k-1}} \rangle$, generating the cognitive state for $p_{ij}$ starts by searching $\mathcal{M}_i$ for $p_{ij}$. If it exists, we retrieve the cognitive state directly; otherwise, we search $\mathcal{M}_i$ for $p_{ij}^{-1}$, i.e., $p_{ij}$ after dropping the last chunk, if exists, we retrieve $s_i^{(i,j_0,\cdots,j_{k-2})}$ from $\mathcal{M}_i$ and use it to generate $s_i^{(i,j_0,\cdots,j_{k-1})}$ as follows: 
\begin{equation*}
    s_i^{(i,j_0,\cdots,j_{k-1})}= \mathcal{A}_i(s_i^{(i,j_0,\cdots,j_{k-2})},c_{j_{k-1}}).
\end{equation*}
The newly generated $s_i^{(i,j_0,\cdots,j_{k-1})}$ is stored in $\mathcal{M}_i$ and indexed by $p_{ij}$.  
This strategy uses a prefix sharing reuse mechanism to convert repeated calculations into lookup operations.


\subsubsection*{Adaptive Pruning}

\noindent Information related to answering a query can be sparsely distributed in a long context. Adaptive pruning dynamically evaluates the value of segmented information and immediately terminates invalid paths. Specifically, if the chunk currently read along a path is deemed useless, the remaining portion of the path is immediately pruned. The interpretation of this pruning strategy is that the order of the chunks in the pruned path makes the information irrational or unenlightened to help answer the question. In this way, the pruning strategy effectively reduces useless path traversal and improves computational efficiency. In Figure \ref{fig:overview}, the red cross under $c_2$ for $\mathcal{A}_0$ illustrates a path pruning. In practice, all cognitive states are still cached normally before the path pruning occurs.



\subsection{Phase 3: Consensus Formation}
The consensus formation phase synthesizes distributed reasoning results through a two-tier hierarchical voting mechanism. 

\subsubsection*{Intra-Agent Aggregation}

\noindent For \(\mathcal{A}_i\) with multiple paths, the answers at the end of each path can be different. In this situation, we prioritize the longest chunk sequence from $\mathcal{M}_i$:
\begin{equation*}
    \sigma_i^* =  \underset{\sigma \in \mathcal{M}_i}{\text{argmax}}|\sigma|,
\end{equation*}
where $\sigma$ is a chunk sequence in the cache $\mathcal{M}_i$. We believe that the longer sequences imply a broader context integration, reducing local bias~\cite{koh2022empirical}.
Each agent then generates its final answer by reprocessing the query with its optimal context:
\begin{equation*}
a_i^{\text{final}} = \mathcal{A}_i \left(q, \mathcal{M}_i(\sigma_i^*), p \right),
\end{equation*}
where $\mathcal{M}_i(\sigma_i^*)$ retrieves the cached state for $\sigma_i^*$.

Note that an agent may return \textbf{none} after reading a path of chunks. It implies that the agent does not find sufficient information to answer the query. 

\subsubsection*{Cross-Agent Majority Voting}

\noindent To resolve agent disagreements, we aggregate all candidates via majority voting:
\begin{equation*}
    a^{\text{final}}=\text{MajorityVote}(\{a_i^{\text{final}}\}_{i=0}^{N-1}).
\end{equation*}
The detailed pseudocode of TOA is provided in Appendix \ref{appendix:alg}.

\section{Experiment Setup}
In this section, we describe the preliminary preparations for our experiments (see Appendix~\ref{appendix:datasets} and \ref{appendix:baselines} for complete setup for datasets and baselines). 

\subsection{Evaluation Datasets}
Experiments are conducted on two long-context reasoning datasets and one benchmark. 

\noindent\textbf{DetectiveQA}~\cite{xu2024detectiveqa}: DetectiveQA is a bilingual dataset with an average question length of 100K, containing a series of detective novel questions and answers. 

\noindent\textbf{NovelQA}~\cite{wangnovelqa}: NovelQA is a dataset with the average input length over 200K for testing the long-text ability of LLMs. It comprises the texts of 89 novels and 2305 question-answer pairs on the details of these novels. 

\begin{figure*}[!t]
  \includegraphics[scale=0.225]{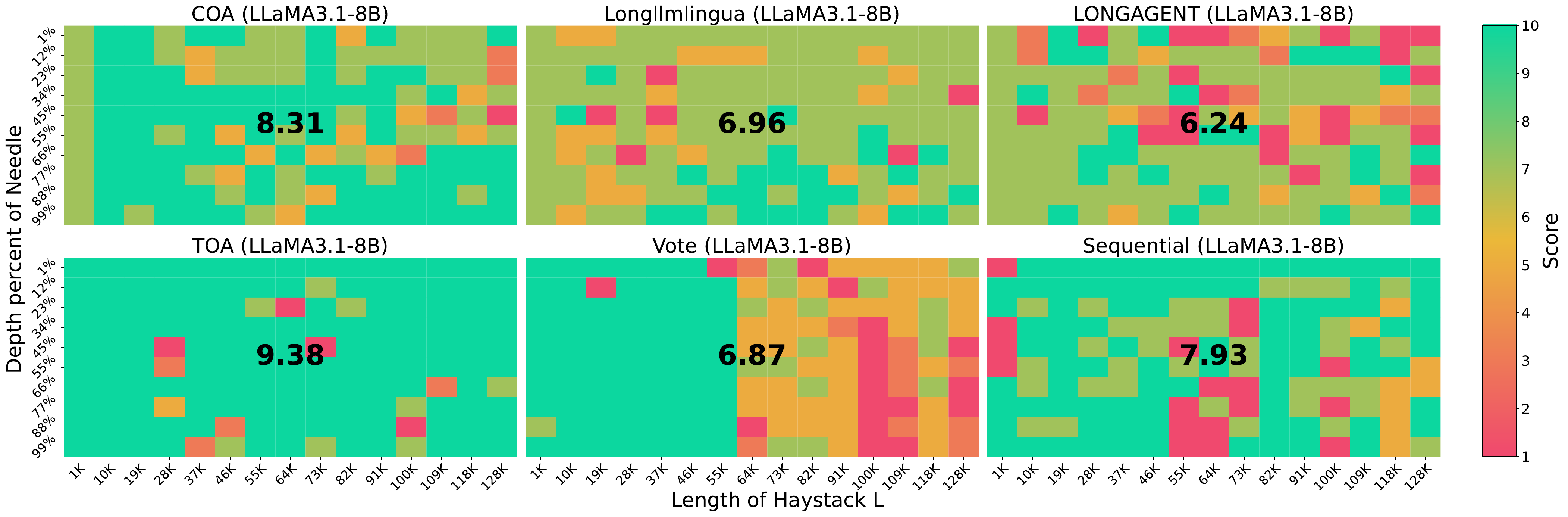} 
  \centering
  \caption {Needle-in-a-Haystack Single-Needle QA results. With TOA, we achieve up to more than 50\% performance improvement compared to the baselines, when the length of haystack changes from 1k to 128k, using the same base model LLama3.1-8B. The percentage value on the y-axis represents the depth percentages of Needle. The bold black numbers in each subfigure indicate the average score.
  }
  \label{fig:single_needle}
\end{figure*}

\begin{figure*}[!h]
  \includegraphics[scale=0.225]{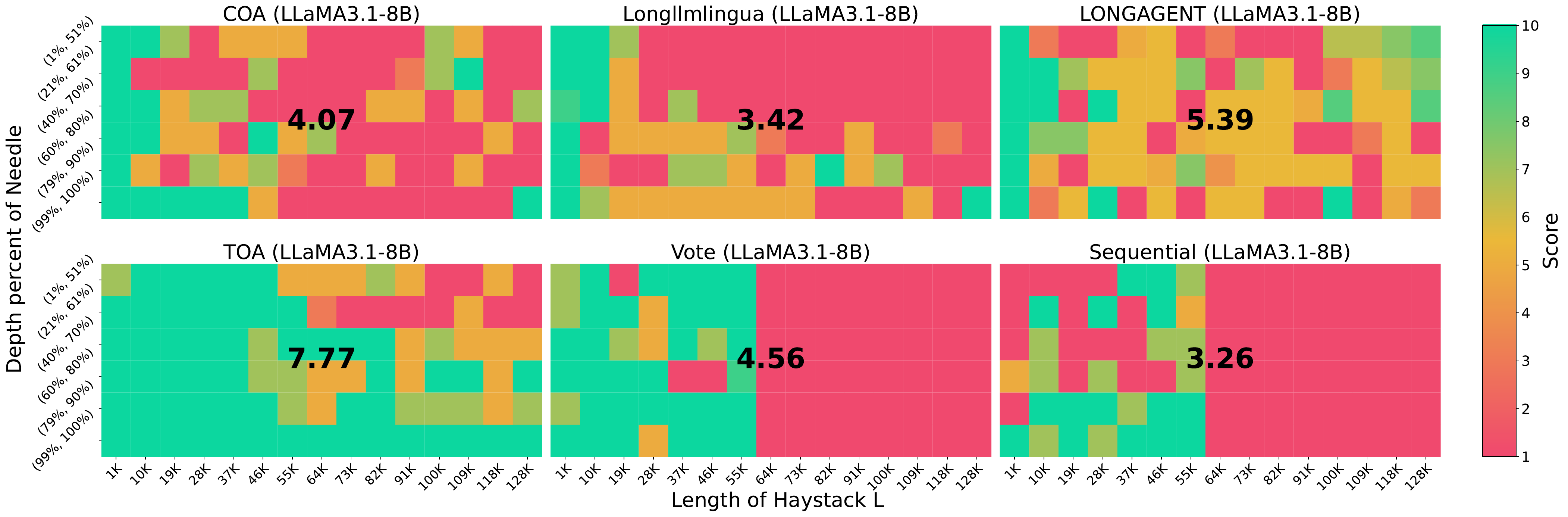} 
  \centering
  \caption {Needle-in-a-Haystack Multi-Needle QA results. With TOA, we achieve up to more than 100\% performance improvement compared to the baselines, when the length of haystack changes from 1k to 128k, using the same base model LLama3.1-8B. The two percentages on the y-axis represent the depth of Needle 1 and Needle 2, respectively. The bold black numbers in each subfigure indicate the average score.}
  \label{fig:multi_needle}
\end{figure*}

\noindent\textbf{Needle-in-a-Haystack}~\cite{kamradt2023needle}: 
\textit{“Needle-in-a-Haystack”} is a widely adopted benchmark for assessing LLMs' ability to process and retrieve critical information from long documents. The method involves inserting one or more short, specific text segments (the needles) into a much longer, semantically unrelated document (the haystack). Adjusting the position of the needle allows us to evaluate the model’s ability to extract key information from various locations in the input. The needle’s depth percentage refers to its insertion point within the haystack. By evaluating the model’s performance at different needle positions, we gain insights into its ability to retrieve information throughout long-context documents. 

\begin{table*}[ht]
\centering
\begin{tabular}{cccccc}
\toprule
                             &                            & \multicolumn{4}{c}{\textbf{Datasets}}                                  \\ \cline{3-6} 
\multirow{2}{*}{\textbf{LLMs}}        & \multirow{2}{*}{\textbf{Baselines}} & \multicolumn{2}{c}{DetectiveQA} & \multicolumn{2}{c}{NovelQA} \\ \cline{3-6} 
                             &                            & Acc$\uparrow$           & None$\downarrow$           & Acc$\uparrow$          & None$\downarrow$        \\ \hline
\multirow{7}{*}{Llama3.1-8B} & COA                        & 0.253 $\pm$ 0.012    & 0.370          & 0.263 $\pm$ 0.017   & 0.160       \\
                             & LONGAGENT                  & \underline{0.487 $\pm$ 0.031}    & 0.157          & 0.373 $\pm$ 0.026   & 0.243       \\
                             & LongLLMLingua              & 0.307 $\pm$ 0.005    & 0.260          & 0.170 $\pm$ 0.000   & 0.500       \\
                             & LongRAG                    & 0.370 $\pm$ 0.000    & 0.217          & \underline{0.440 $\pm$ 0.000}   & 0.153       \\
                             & TOA                        & \textbf{0.543 $\pm$ 0.009}    & \textbf{0.017}          & \textbf{0.450 $\pm$ 0.028}   & \underline{0.043}       \\
                             & Sequential                 & 0.400 $\pm$ 0.028    & 0.143          & 0.257 $\pm$ 0.009   & 0.143       \\
                             & Vote                       & 0.330 $\pm$ 0.022    & \underline{0.023}          & 0.343 $\pm$ 0.017   & \textbf{0.003}       \\ \hline
\multirow{6}{*}{DeepSeek-V3} & COA                        & 0.310 $\pm$ 0.022    & 0.467          & \textbf{0.480 $\pm$ 0.014}   & \textbf{0.110}       \\
                             & LongLLMLingua              & 0.440 $\pm$ 0.043    & 0.257          & 0.410 $\pm$ 0.008   & 0.313       \\
                             & LongRAG                    & 0.420 $\pm$ 0.008    & 0.293          & 0.347 $\pm$ 0.026   & 0.402       \\
                             & TOA                        & \textbf{0.573 $\pm$ 0.005}    & \textbf{0.140}          & \underline{0.473 $\pm$ 0.012}   & \underline{0.120}       \\
                             & Sequential                 & 0.397 $\pm$ 0.025    & 0.360          & 0.423 $\pm$ 0.017   & 0.303       \\
                             & Vote                       & \underline{0.467 $\pm$ 0.019}    & \underline{0.220}          & 0.413 $\pm$ 0.012   & 0.130       \\ \hline
GPT-4o                       & /                          & 0.560 $\pm$ 0.008    & 0.140          & 0.487 $\pm$ 0.012   & 0.270       \\ \hline
Gemini1.5-pro                & /                          & 0.557 $\pm$ 0.005    & 0.090          & 0.457 $\pm$ 0.019   & 0.070       \\ \toprule
\end{tabular}
\caption{Performance comparison on long-context reasoning tasks. We select 100 samples for evaluating each task and repeat it three times to reduce errors. Bold text indicates the best result for the same base model, while an underscore denotes the second-best result. TOA achieves the best performance on most tasks and reaches or even exceeds the performance of much larger commercial models on some tasks with a much smaller base model.}
\label{table:experiment_results}
\end{table*}

\subsection{Evaluation Metrics}
For QA tasks, choosing one from four optional answers, we evaluate performance using two metrics, \textbf{accuracy} and \textbf{none-rate}. \textit{accuracy} refers to the percentage of correct answers. \textit{none} means that the model indicates it cannot retrieve relevant information to answer the question, and \textit{none-rate} is the percentage of \textit{none} in the total test samples, which is an effective reflection of the model's robustness and reliability in open-domain QA tasks. 

For \textit{Needle-in-a-Haystack} tasks, we use GPT-4o (version 2024-05-13) to score (from 1-10) the answers given by our TOA and baselines. A higher score means that the LLM finds more relevant information. 

\subsection{Baselines}
\textbf{LONGAGENT}~\cite{zhao2024longagent}. An approach that uses multi-agent collaboration where the leader breaks down the problem and hands it over to agents for multiple rounds of discussion.

\noindent\textbf{COA}~\cite{zhang2024chain}. CoA consists of multiple worker agents that sequentially communicate to handle different segmented chunks of the text, followed by a manager agent who synthesizes these contributions into a coherent final output. 

\noindent\textbf{LongLLMLingua}~\cite{jiang2024longllmlingua}. A prompt compression method to enhance the perception of LLM for critical information.

\noindent\textbf{LongRAG}~\cite{zhao2024longrag}. A Retrieval Augmented Generation approach to Enhance LLM's understanding of complex contextual knowledge.

\noindent\textbf{Gemini 1.5-pro}~\cite{team2024gemini}. The Gemini 1.5-pro released by Google supports up to 2M context window lengths.

\noindent\textbf{GPT-4o}~\cite{hurst2024gpt}. The GPT-4o model from OpenAI offers a context window of 128K.

\noindent\textbf{Sequential}. The original document is split into multiple chunks, and a single agent reads them sequentially to directly output the final answer.

\noindent\textbf{Vote}. Split the original document into document chunks, then use multiple agents to read the content separately and directly determine the final answer through majority voting.

Note that LongRAG is excluded from the \textit{Needle-in-a-Haystack} test due to the preprocessing overhead brought by building the vector database. LONGAGENT is excluded from the DeepSeek-V3 based experiments because fine-tuning the model is infeasible. 

\subsection{LLMs}
We select LLaMA3.1-8B-instruct~\cite{grattafiori2024llama} and DeepSeek-V3-Chat~\cite{liu2024deepseek} as the base models for TOA and all baselines. The former is deployed locally, while the latter is accessed via API calls.

\begin{table}[]
\centering
\begin{tabular}{lcccc}
\toprule
\multicolumn{1}{c}{\multirow{2}{*}{\textbf{Tasks}}} & \multicolumn{2}{c}{\textbf{DetectiveQA}} & \multicolumn{2}{c}{\textbf{NovelQA}} \\ \cline{2-5} 
                       & Acc$\uparrow$           & None$\downarrow$          & Acc$\uparrow$         & None$\downarrow$        \\ \hline
COA-4K               & 0.253          & 0.370          & 0.263        & 0.160        \\
COA-8K               & 0.340          & 0.290          & 0.350        & 0.030        \\
COA-16K              & 0.450          & 0.180          & 0.410        & \textbf{0.020}        \\
COA-32K              & 0.400          & 0.210          & 0.390        & 0.060        \\ \hline
\multicolumn{1}{c}{TOA-32K}                    & \textbf{0.543}          & \textbf{0.017}          & \textbf{0.450}        & 0.043        \\ \toprule
\end{tabular}
\caption{More comparisons with COA method on long-context reasoning tasks where 4K represents the length of the text chunk assigned to each agent is 4096.}
\label{table:experiment_coa}
\end{table}

\section{Results and discussion}
\subsection{Overall Performance}
\subsection*{Through multi-agent collaboration, TOA may alleviate the lost in the middle problem.} 
Figures \ref{fig:single_needle} and \ref{fig:multi_needle} show that TOA achieves average scores of 9.38 and 7.77 on the single and multi needle tasks, respectively. Notably, its performance remains stable in the middle paragraph range (40\%, 70\%), unlike baselines such as Sequential and Vote, which exhibit significant drops. In the more demanding multi-needle setting, TOA consistently extracts key information, demonstrating superior global alignment across dispersed content.
\subsection*{TOA achieves state-of-the-art performance with lower none-rate.}
Table \ref{table:experiment_results} compares the performance on two long-context reasoning datasets: DetectiveQA and NovelQA, using both LLaMA3.1-8B and DeepSeek-V3 as base models. We observed that COA performed poorly, and we speculate that the problem is that its unidirectional chain suffers from information decay and error amplification issues, which may lead to incomplete or wrong answers. LongLLMLingua greatly compresses the input, but it is easy to lose key information when the query is not directly related to the document, resulting in a high none-rate. LONGAGENT performs relatively well, but it depends on the leader's ability to decompose the problem, and performance may decline when the leader makes a wrong judgment. TOA consistently achieves the highest accuracy across all tasks and models, reaching 54.3\% and 45.0\% on DetectiveQA and NovelQA respectively with LLaMA3.1-8B, and 57.3\% and 47.3\% with DeepSeek-V3. This demonstrates TOA's strong adaptability across different models.

More notably, TOA maintains a remarkably low none-rate—only 1.7\% on DetectiveQA and 4.3\% on NovelQA using LLaMA3.1-8B, suggesting its ability to avoid hallucinated answers in cases of uncertainty. These results support the claim that TOA 
refrains from producing overconfident but incorrect answers.
Interestingly, TOA has a comparable performance against the latest and much larger commercial models, such as Gemini 1.5-pro, by using a much smaller model (LLaMA3.1-8B).
This shows that the improvements stem from architectural innovation rather than scaling alone, and highlights TOA’s parameter efficiency and practical deployment potential. We also investigate the impact of caching and pruning on efficiency, as shown in Appendix~\ref{appendix:efficiency}. 

\begin{figure}[t]
  \includegraphics[scale=0.5]{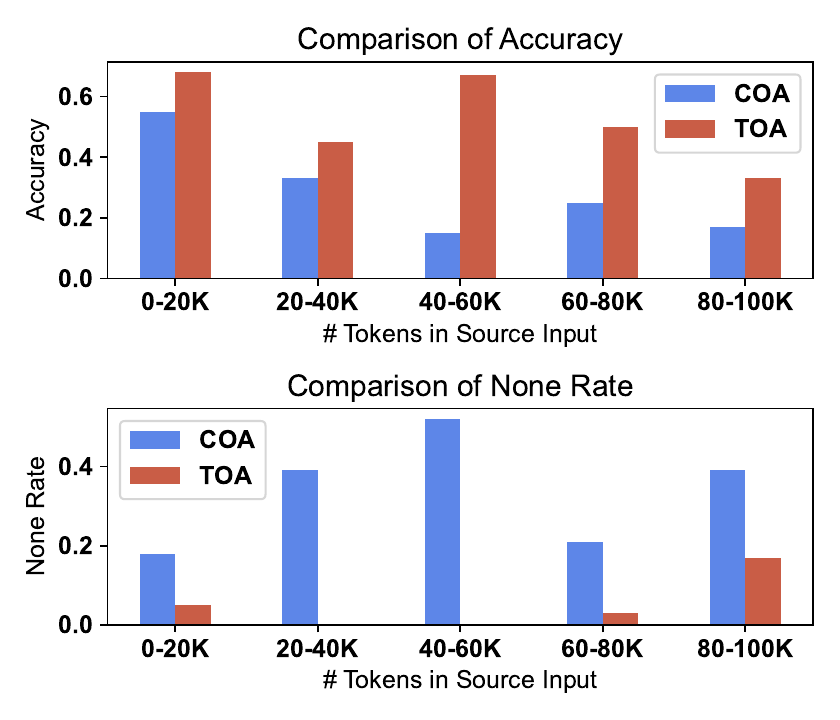} 
  \centering
  \caption {Performance of COA and TOA on DetectiveQA dataset. TOA is more robust to longer inputs.}
  \label{fig:long_context}
\end{figure}

\subsection{Chunk Size Effects}
\label{sec:chunk_size}
We conducted a controlled experiment by varying COA’s chunk size from 4K to 32K—larger chunks imply fewer agents. As shown in Table \ref{table:experiment_coa}, COA’s performance improves with larger chunks from 4K to 16K but drops at 32K. It reflects the limitation of COA’s linear agent chain: deeper information paths increase the risk of context degradation and errors, especially in reasoning tasks. In contrast, TOA allows multi-perspective understanding to support more accurate and stable outputs.

\subsection{Robustness to Input Length}
We further examine how input length affects the performance of TOA. Figure \ref{fig:long_context} shows that TOA remains stable up to 100K tokens, with only a slight accuracy drop and consistently low none-rate. In contrast, COA's performance declines sharply beyond 64K, with rising none-rate. This demonstrates TOA’s robustness to long input. 

\subsection{Impact of Agent Number on Performance}
To explore the impact of agent quantity $N$ on the performance of TOA, we conducted experiments using various agents for QA tasks. The results are shown in Table~\ref{table:agent_performance}.

\begin{table}[]
\centering
\begin{tabular}{ccccc}
\toprule
\multicolumn{1}{c}{\multirow{2}{*}{\textbf{Num}}} & \multicolumn{2}{c}{\textbf{DetectiveQA}} & \multicolumn{2}{c}{\textbf{NovelQA}} \\ \cline{2-5} 
                       & Acc$\uparrow$           & None$\downarrow$          & Acc$\uparrow$         & None$\downarrow$        \\ \hline
Agent=3              & 0.460          & 0.170          & 0.250        & 0.320        \\
Agent=5              & \textbf{0.543}          & \textbf{0.017}          & \textbf{0.450}        & \textbf{0.040}        \\
Agent=7              & 0.510          & 0.040          & 0.280        & 0.190        \\    
\toprule
\end{tabular}
\caption{Performance of TOA with different numbers of agents on DetectiveQA and NovelQA tasks. The accuracy (Acc) should be maximized, and the none rate should be minimized.}
\label{table:agent_performance}
\end{table}

The results show that the performance is optimal when using 5 agents. With fewer agents (3), each agent is tasked with processing larger chunks, which can lead to the "lost in the middle" problem, where key information in the middle of the document is overlooked. On the other hand, when more than 5 agents are used (7 in this case), the text is excessively fragmented, making it harder to synthesize information effectively. Therefore, the use of 5 agents strikes an ideal balance between chunk size and effective information processing, resulting in the best overall performance.

\section{Conclusion}
This study tackles the challenge of long-context reasoning in LLMs. We proposed TOA to prompt multi-perspective understanding of long text by using multiple agents to probe the segmented text chunks in different orders. To enhance computational efficiency, we introduce prefix-hash caching and adaptive pruning techniques. Empirical evaluations show that TOA, despite being built on the lightweight LLaMA3.1-8B, achieves performance on par with significantly larger commercial models on various long-context reasoning tasks.  



\section*{Limitations} 
One limitation of TOA is the additional computational overhead due to probing various orders of text chunks, in particular, as the number of text chunks increases. While caching and pruning techniques are applied, inference remains slower than that of simpler baselines.  This poses a challenge for deployment in large-scale settings. Addressing the scalability will be the focus of our future study.

\cleardoublepage

\appendix
\section{Datasets and Benchmark}
\label{appendix:datasets}
In our experiments, we utilized two datasets: DetectiveQA and NovelQA, selected for their relevance to multi-hop and complex reasoning tasks. For the DetectiveQA dataset, we exclusively used the human-annotated English corpus to ensure high annotation quality. Each question requires the model to select one correct answer from four provided options.  For the NovelQA dataset, we focused on complex reasoning scenarios by specifically extracting multi-hop reasoning questions from the dataset. Due to data leakage concerns, the ground-truth answers for NovelQA are not publicly available. To obtain evaluation results, we submitted the predictions via Codabench~\footnote{\url{https://www.codabench.org/competitions/2727/}}.
Due to computational budget constraints, we randomly sampled 100 examples from each dataset for our experiments. To account for potential model variability, we conducted three independent runs for all experiments and reported the mean value and standard deviation as results. 

We also use the Needle-in-a-Haystack benchmark to assess the ability to retrieve information in long-context documents. In the single-needle setting, we used the following sentence as the needle: "The production company for \textit{The Year Without a Santa Claus} is best known for seasonal television specials, particularly its work in stop-motion animation." The corresponding question was: "For what type of work is the production company for \textit{The Year Without a Santa Claus} best known?". 
In the multi-needle setting, we used two distinct needle sentences:

\begin{itemize}
    \item ``According to declassified Cold War documents, spies used a hollowed-out chess piece as a dead drop in 1970s Berlin.''
    \item ``According to declassified Cold War documents, a fake electrical fuse box was used as a dead drop by spies in 1970s Berlin.''
\end{itemize}
The associated question was: ``According to declassified Cold War documents, what were the two unusual objects that spies used as dead drops in 1970s Berlin?'' 
    

\section{Baselines} 
\label{appendix:baselines}
All experiments were conducted using a machine equipped with two NVIDIA GeForce RTX 3090 GPUs, an Intel Core i9-14900K CPU, and 128 GB of RAM. Implementations were based on PyTorch v2.6.0. To ensure consistency across all baselines, we set the decoding temperature to 0.01 and the maximum output length to 2048 tokens. For locally run models (e.g., LLaMA3.1-8B-instruct), we used FP16 precision and performed inference on the same hardware setup. 

\begin{itemize}
\item For LONGAGENT, we manually reimplemented the method by closely following the steps described in the original paper. Specifically, we fine-tuned LLaMA3.1-8B on 10,000 positive and 15,000 negative samples (constructed from the same data distribution). The chunk size was set to the default value of 4096 tokens.

\item For COA, we used its open-source implementation with the default chunk size of 4096. To evaluate the impact of different chunk sizes, apart from the default size of 4096, we also tested COA on various chunk sizes (see Section~\ref{sec:chunk_size} for details). 

\item For LongLLMLingua, we used the open-source prompt compression implementation with all parameters at their default values. The compressed prompts were subsequently passed to our base model for evaluation. 

\item For LongRAG, we employed the official open-source implementation. The chunk size was set to the maximum supported value of 500, with all other parameters left at default. Prior to evaluation, all documents were preprocessed via chunking and vectorization for retrieval.

\item For commercial models such as Gemini-1.5-Pro and GPT-4o, we directly fed the documents and questions into the models and prompted them to generate answers without further customization.

\item For TOA, we set the number of agents to 5. In both TOA and the Vote baseline, if some agents returned 'None' during the voting process, we only considered valid responses. In cases of a tie, an additional independent agent was queried to make the final decision. 
\end{itemize}

\section{The Impact of Reading Order}
\label{appendix:order}
To better understand the importance of reading order in multi-hop reasoning, we present a simple illustrative example:\\

Imagine a short story:  
\begin{enumerate}
    \item "The room was messy."  
    \item "John sighed."  
    \item "He had just finished a big project." 
\end{enumerate}

If you read these sentences in order (1→2→3), you might infer that John sighed because the messy room stressed him out. However, if you reverse the order (3→2→1), it suggests that John sighed in relief after completing his project, and the messy room is merely a background detail. The same facts, presented in different orders, lead to entirely different interpretations. By probing various orders of text chunks, TOA realizes multi-perspective reasoning and avoids being constrained to a single narrative flow. This enables us to uncover how order shapes meaning, potentially help us reveal hidden biases or overlooked perspectives. 

\section{Efficiency}
\label{appendix:efficiency}
To better measure the efficiency of our method, Table~\ref{tab:efficiency} presents the number of API calls by TOA and baseline COA. Both use DeepSeek-V3 as the base model on the NovelQA dataset. The main efficiency gains in TOA occur in phase 2, where caching and pruning are applied. Without caching and pruning, phase 2 requires an estimated 2103 API calls. Enabling caching alone reduces this number to 1830, saving 273 calls (a 13.0\% reduction). When both caching and pruning are applied, the number of calls drops further to 1034 — saving at least 1069 calls (a 50.8\% improvement). 

We further compared the number of token requests in Table~\ref{tab:token_savings}. Pruning and caching significantly reduced overhead. It's worth noting that this mechanism relies on the capabilities of the base model itself, so its effect is not significant on models with small parameters.
Although the total cost in TOA is slightly higher than that of COA, this additional cost yields a notable improvement in output. 

\begin{table*}[h]
\centering
\begin{tabular}{lllccc}
\toprule
\textbf{Method} & \textbf{Phase} & \textbf{Strategy} & \textbf{API Calls} & \textbf{Saved Calls} & \textbf{Saving Rate} \\
\midrule
\multirow{5}{*}{\textbf{TOA}} 
  & \multirow{3}{*}{Phase 2} 
    & w/o Caching \& Pruning     & 2103 & --       & -- \\
  &  
    & w/ Caching Only            & 1830 & 273      & 13.0\% \\
  &  
    & w/ Caching \& Pruning      & 1034 & 1069     & 50.8\% \\
\cmidrule(lr){2-6}
  & Phase 1 \& 3 
    & --            & 1500 & --       & -- \\
\cmidrule(lr){2-6}
  & All Phases 
    & w/ Caching \& Pruning      & \textbf{2534} & -- & -- \\
\midrule
\textbf{COA} & -- & -- & \textbf{2287} & -- & -- \\
\bottomrule
\end{tabular}
\caption{The number of API calls by TOA and COA on the NovelQA dataset. The evaluation is conducted on the same 100 examples used in the experiments.} 
\label{tab:efficiency}
\end{table*}

\begin{table*}[h]
\centering
\small
\begin{tabular}{ccccccc}
\toprule
\textbf{Dataset} & \textbf{Model} & \multicolumn{3}{c}{\textbf{TOA Tokens}} & \textbf{Token Savings} & \textbf{COA Tokens} \\
\cmidrule{3-5}
 & & NO Cache\&Prune & with Cache & with Cache+Prune & & \\
\midrule
\multirow{2}{*}{NovelQA} 
 & DeepSeek-V3 & 31210K & 25609K & 12804K & 59\% & 10229K \\
 & Llama3.1-8B & 24798K & 23495K & 23476K & 5\% & 10229K \\
\midrule
\multirow{2}{*}{DetectiveQA} 
 & DeepSeek-V3 & 18438K & 15522K & 12292K & 33\% & 5414K \\
 & Llama3.1-8B & 12575K & 11933K & 11932K & 5\% & 5414K \\
\bottomrule
\end{tabular}
\caption{Token statistics across datasets and models with different caching/pruning settings. The evaluation is conducted on the same 100 examples used in the experiments.}
\label{tab:token_savings}
\end{table*}

\section{Algorithm for TOA}
\label{appendix:alg}
The pseudocode of TOA is presented in Algorithm~\ref{alg:alg1}. 

\begin{algorithm*}[!ht]
    \renewcommand{\algorithmicrequire}{\textbf{Input:}}
    \renewcommand{\algorithmicensure}{\textbf{Output:}}
    \caption{TOA algorithm.}
    \label{alg:toa}
    \begin{algorithmic}[1]
        \REQUIRE Document $\mathcal{D}$, chunk number $N$, query $q$ and prompt $p$.
        \ENSURE Final answer $a^{\text{final}}$.
        
        \STATE Split $\mathcal{D}$ into small chunks $\mathcal{C}=\{c_i\}^N_{i=0}$
        \FOR{$i = 0$ to $N-1$}
            \STATE Initialize $\mathcal{M}_i$ to store all cognitions
            \STATE Initialize $\mathcal{U}_i$ to store all chunk sequence usefulness
            \STATE $s_i^{(i)} \gets \langle e_i^{(i)}, a_i^{(i)} \rangle = \mathcal{A}_i(q, c_i, p)$
            \STATE $\mathcal{M}_i(\langle c_i \rangle) \gets s_i^{(i)}$
        \ENDFOR
        
        \FOR{$i = 0$ to $N-1$}
            \STATE Get the helpful chunks index set $\mathcal{G}_i$
            \IF{$\mathcal{G}_i \neq \emptyset$}
                \STATE $\text{Perm}_i \gets \text{Permutations}(\mathcal{G}_i)$
                \FOR{$\pi \in \text{Perm}_i$}
                    \STATE $\mathcal{P}_{\pi} = \langle c_i \rangle \oplus \langle c_{\pi_1}, c_{\pi_2}, ..., c_{\pi_k}\rangle$
                    \FOR{$r = 2$ to $|\mathcal{P}_{\pi}|$}
                        \STATE $\sigma_r \gets \mathcal{P}_\pi[0:r]$ 
                        \IF{$\sigma_r\in \mathcal{U}_i$}
                            \IF{$\mathcal{U}_i(\sigma_r)==0$}
                                \STATE \textbf{break}
                            \ELSE                            \STATE $s_i^{\sigma_r}=\mathcal{M}_i(\sigma_r)$
                            \STATE \textbf{continue}
                            \ENDIF
                        \ENDIF
                        \IF{$\text{Utility}(s_i^{\sigma_{r-1}},c_{\pi_r})==0$}
                            \STATE $\mathcal{U}_i(\sigma_r)\gets0$
                            \STATE \textbf{break}
                        \ELSE
                            \STATE $s_i^{\sigma_r}\gets \mathcal{A}_i(s_i^{\sigma_{r-1}},c_{\pi_r})$
                            \STATE $\mathcal{M}_i(\sigma_r) \gets s_i^{\sigma_r}$                       \STATE $\mathcal{U}_i(\sigma_r) \gets 1$ 
                            
                        \ENDIF
                    \ENDFOR
                \ENDFOR
            \ENDIF
        \ENDFOR
        \FOR{$i = 0$ to $N-1$}
            \STATE $\sigma_i^* \gets \text{argmax}_{\sigma \in (\mathcal{M}_i)} |\sigma|$ 
            \STATE $a_i^{\text{final}} \gets \mathcal{A}_i(q, \mathcal{M}_i(\sigma_i^*), p)$ 
        \ENDFOR
        \STATE $a^{\text{final}} \gets \text{MajorityVote}(\{a_i^{\text{final}}\}_{i=0}^{N-1})$ 
        \RETURN $a^{\text{final}}$
    \end{algorithmic}
    \label{alg:alg1}
\end{algorithm*}


\section{Prompt Templates Used in TOA}
Table~\ref{tab:satge1} - \ref{tab:tiebreaker} present the prompt templates used in the three phases of TOA, respectively. 

\begin{table*}[h]
\centering
\begin{tabular}{p{0.95\linewidth}}
\toprule
\textbf{Prompt Template for Chunk Perception (Phase 1)} \\
\midrule
\textbf{Background:} \\
You are a skilled agent tasked with answering a question based on a long context. Since the context is too long, it is divided into chunks, each assigned to a different agent. \\
\\
\textbf{Task:} \\
You are in Phase 1. Given a document chunk and a multiple-choice question, your goal is to answer the question accurately. First, extract and summarize facts relevant to the question from your assigned segment. Then, draw your conclusion based solely on those facts. Do not rely on prior knowledge. \\
\\
\textbf{Output Format (JSON):} \\
\texttt{\{ } \\
\texttt{\ \ \ "evidence": "Factual excerpts supporting your reasoning",} \\
\texttt{\ \ \ "answer": "Your answer based on the evidence"} \\
\texttt{\}} \\
\bottomrule
\end{tabular}
\caption{Prompt template (Phase 1).}
\label{tab:satge1}
\end{table*}

\begin{table*}[h]
\centering
\begin{tabular}{p{0.95\linewidth}}
\toprule
\textbf{Prompt Template for Multi-Perspective Understanding (Phase 2-1)} \\
\midrule
\textbf{Background:} \\
You are a skilled agent tasked with answering a question based on a document. Since the document is too long, it is divided into multiple chunks, each read by a different agent. \\
\\
\textbf{Task:} \\
You are in Phase 2-1. You have already read your assigned chunk and proposed an evidence-based answer. However, your view may be incomplete or incorrect due to the limited context. \\
\\
You will now be shown the evidence and answers provided by other agents who read different chunk of the document. The correct answer may appear in one or more of these responses. \\
\\
\textbf{Note:} \\
If only a few agents report relevant evidence while most say there is none, you should focus on those few with relevant content. \\
\\
\textbf{Decision:} \\
Select which agent(s)’ responses may help refine your understanding without introducing irrelevant information. You may choose one or more agent IDs, or \texttt{"None"} if no agent adds value. \\
\\
Use only the information shown. Do not use external knowledge. \\
\\
Valid choices: \texttt{\{agent\_list\}} \\
\\
\textbf{Output Format (JSON):} \\
\texttt{\{} \\
\texttt{\ \ "explanation": "Justify your selection.",} \\
\texttt{\ \ "id": "Selected agent ID(s), e.g., '0', '0,1', or 'None'"} \\
\texttt{\}} \\
\bottomrule
\end{tabular}
\caption{Prompt template (Phase 2-1).}
\label{tab:stage2-1}
\end{table*}

\begin{table*}[h]
\centering
\begin{tabular}{p{0.95\linewidth}}
\toprule
\textbf{Prompt Template for Multi-Perspective Understanding (Phase 2-2)} \\
\midrule
\textbf{Background:} \\
You are a skilled agent tasked with answering a question based on a document. Since the document is too long, it is divided into chunks, each read by a different agent. \\
\\
\textbf{Task:} \\
You are in Phase 2-2. Based on your earlier reasoning, you requested to view additional text chunks from other agents to refine your understanding. \\
\\
You will now be shown one of these chunks. Carefully evaluate its relevance. If the chunk only repeats known information or introduces irrelevant content, mark it as \texttt{"useless"}. Otherwise, mark it as \texttt{"useful"} and update your facts and conclusion accordingly. \\
\\
If the chunk is \texttt{"useless"}, repeat your original facts and conclusion. Limit your output length—abbreviate if necessary. \\
\\
You must judge only based on the content provided, not using external knowledge. \\
\\
\textbf{Output Format (JSON):} \\
\texttt{\{} \\
\texttt{\ \ "utility": "useless" or "useful",} \\
\texttt{\ \ "fact": "Updated factual summary.",} \\
\texttt{\ \ "conclusion": "Updated answer based on new information."} \\
\texttt{\}} \\
\bottomrule
\end{tabular}
\caption{Prompt template (Phase 2-2).} 
\label{tab:stage2-2}
\end{table*}

\begin{table*}[h]
\centering
\begin{tabular}{p{0.95\linewidth}}
\toprule
\textbf{Prompt Template for Consensus Formation (Phase 3-1)} \\
\midrule
\textbf{Background:} \\
You are a skilled agent tasked with answering a question based on a document. Since the document is too long, it is divided into chunks, each read by a different agent. \\
\\
\textbf{Task:} \\
You are in the final phase. All agents have now exchanged their opinions. Based solely on the question, answer options, and your aggregated opinion, provide the final answer. \\
\\
If you are still uncertain and unable to choose a valid option, respond with \texttt{"None"}. \\
\\
\textbf{Note:} \\
All relevant information has been condensed into your own opinions. Do not consider external content or reprocess the original document. Make your decision based only on your internal conclusion. \\
\\
\textbf{Output Format (JSON):} \\
\texttt{\{} \\
\texttt{\ \ "explanation": "Brief reasoning for your choice.",} \\
\texttt{\ \ "result": "One of A, B, C, D, or None (no punctuation)"} \\
\texttt{\}} \\
\bottomrule
\end{tabular}
\caption{Prompt template (Phase 3-1).}
\label{tab:stage3}
\end{table*}

\begin{table*}[h]
\centering
\begin{tabular}{p{0.95\linewidth}}
\toprule
\textbf{Prompt Template for Tie-Breaking Decision (Phase 3-2)} \\
\midrule
\textbf{Background:} \\
You are the final decision maker. You are presented with a long document and a multiple-choice question. \\
\\
There are \texttt{\{len(Agent\_list)\}} decision makers. A majority vote was attempted, but a tie occurred. \\
\\
\textbf{Task:} \\
Please examine each agent's factual conclusions and opinions carefully. Based on this information, select the best final answer. \\
\\
\textbf{Rules:} \\
1. You \textbf{MUST} choose from the following options: \texttt{\{result\}} \\
2. \textbf{DO NOT} generate any answer outside this list. \\
3. Output your decision strictly in the following JSON format. \\
\\
\textbf{Tie Information:} \\
Answers with the same number of votes: \texttt{\{result\}} \\
\\
\textbf{Output Format (JSON):} \\
\texttt{\{} \\
\texttt{\ \ "explanation": "Justify your choice.",} \\
\texttt{\ \ "result": "Final answer choice, e.g., A or B"} \\
\texttt{\}} \\
\bottomrule
\end{tabular}
\caption{Prompt template (Phase 3-2).}
\label{tab:tiebreaker}
\end{table*}

\section{Case Study}
A case study is described in Table~\ref{tab:case-study-q18-full}. 

\subsection*{Phase 1: Chunk Perception}

The document is split into five chunks and allocated to five agents, each for one. Each agent generates the first response based on their chunks. Because on different chunks, agents can respond vary differently. For instance, Agent 0 identifies the victims as hotel regulars, whereas Agent 1 views them as business associates. 

\subsection*{Phase 2: Multi-Perspective Understanding}

Once the agents generate their initial cognition, they proceed to the multi-perspective understanding. Agents exchange their cognition and request access to other agents’ chunks to update their cognition. A key feature is that each agent probes the different orders of text chunks for multi-perspective understanding of the input document. For example, Agent 0 explores text chunks 2, 3, and 4. During this process, agents evaluates the utility of each chunk. If a chunk is deemed unhelpful (marked as "useless"), the path is pruned. The pruning strategy ensures that reasoning efforts are focused on informative content, reducing unnecessary computation. Another strategy applied is state caching with prefix hashing. For instance, Agent 0 reuses the cached state after evaluating Agent 3’s chunk. This caching strategy is especially effective when certain sequences of chunks are revisited when probing paths.


\subsection*{Phase 3: Consensus Formation}


Following multi-perspective understanding, agents synthesize the gathered information to form a final consensus. In this example, all five agents converge on the same answer \textbf{A}. The final decision is determined through majority vote among agents, targeting for a more robust and collectively reasoned decision. 

\newpage
\begin{table*}[h]
\small
\centering
\renewcommand{\arraystretch}{1.2}
\begin{tabularx}{\textwidth}{l X}
\toprule
\textbf{Query} &
\textit{"What is the relationship between the three victims?"} \\
\midrule
\textbf{Assign Documents} &
\texttt{Assigned to 5 agents, length 5*4428.0 = 22140} \\
\midrule
\textbf{Chunk Perception} &
\textbf{Agent 0:} D.  
\quad \textit{They were all regular customers at the same hotel.} \newline
\textbf{Agent 1:} C.  
\quad \textit{The victims were known to frequently exchange business proposals.} \newline
\textbf{Agent 2:} C.  
\quad \textit{All three victims had booked rooms under the same group reservation. } \newline
\textbf{Agent 3:} C.  
\quad \textit{Each victim was connected to a similar project involving a large sum of money. } \newline
\textbf{Agent 4:} A.  
\quad \textit{The victims had a shared history of working together. } \\
\midrule
\textbf{Multi-Perspective Understanding} &
\textbf{Agent 0:} I saw Agent [1, 2, 3, 4]'s cognition, want to see Agent [2, 3, 4]'s chunk. \newline
\textbf{Agent 1:} I saw Agent [0, 2, 3, 4]'s cognition, want to see Agent [4]'s chunk. \newline
\textbf{Agent 2:} I saw Agent [0, 1, 3, 4]'s cognition, want to see Agent [4]'s chunk. \newline
\textbf{Agent 3:} I saw Agent [0, 1, 2, 4]'s cognition, want to see Agent [4]'s chunk. \newline
\textbf{Agent 4:} I saw Agent [0, 1, 2, 3]'s cognition, want to see Agent [0]'s chunk. \\
\midrule
\textbf{Multi-Perspective Understanding} &
\textbf{Agent 0:} \newline
\quad Begin sequence [2, 3, 4].\newline
\quad ---Saw Agent [2]'s chunk — \texttt{useless}, marked (0, 2) as useless.\newline
\quad ------Because [2] is useless, skip the rest of the sequence.\newline
\quad Begin sequence [2, 4, 3].\newline
\quad ---(0, 2) is already proved useless, skip the rest of the sequence.\newline
\quad Begin sequence [3, 2, 4].\newline
\quad ---Saw Agent [3]'s chunk - \texttt{useful}.\newline
\quad ------New cognition added: (0, 3).  \newline
\quad ---Saw Agent [2]'s chunk - \texttt{useless}, marked (0, 3, 2) as useless.\newline
\quad ------Because [2] is useless, skip the rest of the sequence.\newline
\quad Begin sequence [3, 4, 2].\newline
\quad ---Load cache (0, 3), Saw Agent [4]'s chunk - \texttt{useful}.\newline
\quad ------New cognition added: (0, 3, 4).\newline
\quad ---Saw Agent [2]'s chunk - \texttt{useless}, marked (0, 3, 4, 2) as useless.\newline
\quad Begin sequence [4, 2, 3].\newline
\quad ---Saw Agent [4]'s chunk - \texttt{useful}.\newline
\quad ------New cognition added: (0, 4).\newline
\quad ---Saw Agent [2]'s chunk - \texttt{useless}, marked (0, 4, 2) as useless.\newline
\quad ------Because [2] is useless, skip the rest of the sequence. \newline
\quad Begin sequence [4, 3, 2].\newline
\quad ---Load cache (0, 4), Saw Agent [3]'s chunk - \texttt{useful}.\newline
\quad ------New cognition added: (0, 4, 3).\newline
\quad ---Saw Agent [2]'s chunk - \texttt{useful}.\newline
\quad ------New cognition added: (0, 4, 3, 2).\newline

\textbf{Agent 1:} \newline
\quad Begin Sequence [4].\newline
\quad ---Saw Agent [4]'s chunk — \texttt{useful} \newline
\quad ------New cognition added: (1, 4) \newline

\textbf{Agent 2:} \newline
\quad Begin Sequence [4].\newline
\quad ---Saw Agent [4]'s chunk — \texttt{useful} \newline
\quad ------New cognition added: (2, 4) \newline

\textbf{Agent 3:} \newline
\quad Begin Sequence [4].\newline
\quad ---Saw Agent [4]'s chunk — \texttt{useful} \newline
\quad ------New cognition added: (3, 4) \newline

\textbf{Agent 4:} \newline
\quad Begin Sequence [0].\newline
\quad ---Saw Agent [0]'s chunk — \texttt{useful} \newline
\quad ------New cognition added: (4, 0) \\
\midrule
\textbf{Cognition Summary} &
(0), (0, 3), (0, 4), (0, 3, 4), (0, 4, 3), (0, 4, 3, 2) \quad (1), (1, 4) \quad (2), (2, 4) \quad (3), (3, 4) \quad (4), (4, 0) \\
\midrule
\textbf{Longest Cognitions} &
(0, 4, 3, 2),\quad (1, 4), \quad(2, 4), \quad(3, 4),\quad (4, 0)\\
\midrule
\textbf{Consensus Formation} &
\textbf{Agent 0:} A \quad
\textbf{Agent 1:} A \quad
\textbf{Agent 2:} A \quad
\textbf{Agent 3:} A \quad
\textbf{Agent 4:} A \newline
\textbf{Majority Answer:} \texttt{A} \\
\bottomrule
\end{tabularx}
\caption{A case study}
\label{tab:case-study-q18-full}
\end{table*}

\end{document}